\def\year{2019}\relax
\documentclass[letterpaper]{article} %DO NOT CHANGE THIS
\usepackage{aaai19}  %Required
\usepackage{times}  %Required
\usepackage{helvet}  %Required
\usepackage{courier}  %Required
\usepackage{url}  %Required
\usepackage{graphicx}  %Required
\usepackage{amsmath}
\usepackage{amssymb}
\usepackage{prettyref}

\begin{document}
 	\newcommand{\mvec}[1]{\mathbf{#1}}
 	\newcommand{\mset}[1]{\mathbf{#1}}
 	\newcommand{\mmat}[1]{\mathbf{#1}}
 	\newcommand{\argmax}{\mathop{\rm argmax}\limits}
 	\newcommand{\argmin}{\mathop{\rm argmin}\limits}
 	\newcommand{\commentsize}{\footnotesize}
 	\newcommand{\Pre}{\mathbf{\rm Pre}}
 	\newcommand{\sitepath}[1]{{\sf #1}}
 	\newcommand{\action}[1]{{\tt #1}}
 	\newcommand{\nonterminal}[1]{{\sf #1}}
 	\newcommand{\method}[1]{{\sf #1}}
 	
 	\newcommand{\defeq}{\ensuremath{\stackrel{\mathrm{def}}{=}}}
 	
 	\newcommand{\db}{\mathit{DB}}
 	\newcommand{\pdb}{P_\mathit{DB}}
 	\newcommand{\msw}{\mbox{\tt msw}}
 	\newcommand{\tensor}{\mbox{\tt tensor}}
 	\newcommand{\indexfunc}{T}
 	\newcommand{\einsum}{\mbox{einsum}}
 	\newcommand{\minx}{\mbox{min}_1}

 	\newrefformat{fig}{Figure \ref{#1}}
 	\newrefformat{tbl}{Table \ref{#1}}
 	\newrefformat{eq}{Equation (\ref{#1})}
 	
%\title{T-PRISM: integration with continuous embeddings \\ and a probabilistic logic programming language}
%\title{Tensorized logic programming for modeling by T-PRISM}
%\title{High-dimensional and large-scale data modeling using tensorized logic programming by T-PRISM}
\title{A tensorized logic programming language for large-scale data}
%\title{Tensorized probabilistic logic programming by T-PRISM}
\author{Ryosuke Kojima$^{1}$, Taisuke Sato$^{2}$\\
$^{1}$ Department of Biomedical Data Intelligence, Graduate School of Medicine,\\
Kyoto University, Kyoto, Japan.\\
$^{2}$ AI research center (AIRC)\\
National Institute of Advanced Industrial Science and Technology (AIST), Tokyo, Japan.
}

\maketitle
\begin{abstract}
We introduce a new logic  programming language T-PRISM based on tensor
embeddings.    Our  embedding   scheme  is   a  modification   of  the
distribution  semantics   in  PRISM,   one  of   the  state-of-the-art
probabilistic logic  programming languages, by  replacing distribution
functions  with  multi-dimensional  arrays,  i.e.,  tensors.   T-PRISM
consists  of   two  parts:   logic  programming  part   and  numerical
computation  part.  The  former  provides  flexible and  interpretable
modeling  at the  level  of first  order logic,  and  the latter  part
provides scalable  computation utilizing parallelization  and hardware
acceleration with GPUs.  Combing these two parts provides a remarkably
wide range of high-level  declarative modeling from symbolic reasoning
to deep learning.

%in a unified way  where we express entities  and relations at the  level of
%first order logic  but learn and evaluate them in  a vector space.

To  embody  this  programming  language,   we  also  introduce  a  new
semantics, termed tensorized semantics, which combines the traditional
least  model semantics  in logic  programming with  the embeddings  of
tensors.  In  T-PRISM, we first derive  a set of equations  related to
tensors from  a given  program using  logical inference,  i.e., Prolog
execution in a symbolic space and  then solve the derived equations in
a continuous space by TensorFlow.

Using our preliminary implementation  of T-PRISM, we have successfully
dealt  with  a wide  range  of  modeling.
We have
succeeded in  dealing with  real large-scale  data in  the declarative
modeling.   This paper  presents a DistMult
model for knowledge graphs using the  FB15k and WN18 datasets.
\end{abstract}

\section{1. Introduction}
\noindent 
%Integrating symbolic reasoning  and low-level  noisy data  processing
%such  as image  recognition  and  signal processing  has  long been  a
%challenging problem in real-world applications of AI.
%Recently, to handle a wide range of applications
%for high dimensional and large-scale data, the unification of neural
%networks and approximate reasoning by symbol embeddings into
%continuous spaces has been proposed %\cite{manhaeve2018deepproblog,rocktaschel2017end,evans2018learning}.
%Also in the different context, 
%the framework of logic programming, the integration of logic and
%probability has been extensively studied as probabilistic logic programming
%(PLP) \cite{kimmig2011implementation,wang2013programming,Sato08a}.

Logic programming provides concise expressions of knowledge and has
been proposed as means for representing and modeling various types of
data for real-world AI systems.  For example, to deal with uncertain
and noisy data, probabilistic logic programming (PLP) has been
extensively
studied \cite{kimmig2011implementation,wang2013programming,Sato08a}.
PLP systems allow users to flexibly and clearly describe stochastic
dependencies and relations between entities using logic programming.
Also in the different context, to handle a wide range of applications,
the unification of neural networks and approximate reasoning by symbol
embeddings into continuous spaces has recently been
proposed \cite{manhaeve2018deepproblog,rocktaschel2017end,evans2018learning}.
%which can take advantage of hardware acceleration for processing large
%scale data.
%kok2010learning

In this  paper, we tackle a  task of combining symbolic  reasoning and
multi-dimensional  continuous-space embeddings  of logical  constructs
such  as clauses,  and explore  a new  approach to  compile a  program
written  in  a declarative  language  into  a procedure  of  numerical
calculation  suitable  for  large-scale   data.   Such  languages  are
expected  to   be  interpretable  as  a   programming  language  while
efficiently  executable in  the  level of  numerical calculation  like
vector computation.  Aiming at this  goal, we introduce {\em tensorized
semantics\/},  a  novel  algebraic semantics  interfacing  a  symbolic
reasoning layer and  the numeric computation layer, and  propose a new
modeling   language  ``T-PRISM''.    It  is   based  on   an  existing
probabilistic  logic  programming  language PRISM  \cite{Sato08a}  and
implements the tensorized  semantics for large-scale datasets.

Thus the first contribution of this paper is the introduction of a new
semantics, {\em  tensorized semantics\/}.   The current PRISM  has the
distribution    semantics    \cite{Sato95}   that    probabilistically
generalizes  the  least  model   semantics  in  logic  programming  to
coherently assign  probabilities to the logical  constructs.  Likewise
tensorized semantics assigns tensors\footnote{
In  this paper,  the term  ``tensor'' is  used as  a multi-dimensional
array interchangeably.
} to the logical constructs based  on the least model semantics.  Both
of PRISM and T-PRISM programs are  characterized by a set of equations
for the assigned quantities.

One may be  able to view T-PRISM as one  approach to an equation-level
interface   to   connect   logic  programming   and   continuous-space
embeddings.  Another approach  is possible, predicate-level interface,
such as DeepProblog \cite{manhaeve2018deepproblog}.  Their approach is
to  provide   special  predicates   connecting  neural   networks  and
probabilistic  logic programming  by  ProbLog.  It  implementationally
separates neural networks from  probabilistic models but syntactically
integrates, for example, image recognition and logical reasoning using
estimated labels.  This approach, unlike  our approach, does not allow
constants  and  predicates  to have  corresponding  vectors  (tensors)
representations in neural networks.

The second contribution of this paper is an implementation methodology
of the T-PRISM's tensorized semantics.  T-PRISM's new semantics guides
an internal  data structure  to connect  symbolic inference  by Prolog
execution         and         numerical         calculation         by
TensorFlow \cite{abadi2016tensorflow}.
This internal  data structure
represents tensor equations for tensorized semantics
and is an expanded PRISM's explanation graph, {\em explanation  graphs  with Einstein's  notation\/},  a set  of
logical equations  (bi-implications) that allows  repeated sum-product
operations  for  a  specified   number  of  times,  called  Einstein's
notation.
This expansion is intended
for  parallelization  and  enhance  the applicability  of  T-PRISM  to
large-scale data.
%Tabling

%In the following, we first describe  the semantics of T-PRISM and then
%its implementation  using explanation graphs expanded  for large-scale
%data and  a wide  range of  modeling.  After  that, we  explain sample
%programs and detail  two experiments with real  datasets.  Finally, we
%summarize this paper.

\section{2. Tensorized semantics}

T-PRISM's   semantics,  ``tensorized   least   model  semantics''   or
``tensorized  semantics''  for  short,  is  a  tensorized  version  of
existing PRISM semantics.  In PRISM, the semantics assigns probability
masses to  ground atoms in the  Herbrand base of a  program whereas in
T-PRISM, multi-dimensional arrays are assigned to them.

Formally, a T-PRISM   program    is   a   five-tuple   $\left\langle
F,C,q_F,\indexfunc_F,R\right\rangle$.   A  primary   part  of  the
program $\db=F  \cup C $ is  a Prolog (definite clause)  program where
$F$ is a set of {\em tensor atom\/}s  (see below), and $C$ is a set of
definite clauses  whose head contains no  tensor atom.  Theoretically,
we  always  equate  a set  of  clauses  with  the  set of  all  ground
instances and allow $\db$ to be countably infinite.

In addition  to this setting,  we introduce  a specific type  of atoms
called  {\em  tensor  atom\/}s  represented by  {\tt  tensor/2}  which
declare multi-dimensional  arrays called {\em embedded  tensor\/}s.  A
tensor atom has an {\em index\/} as  a list of Prolog constants in the
second   argument   to   access    the   entries   of   the   declared
multi-dimensional   array.   For   example,  ${\tt   tensor(x,[i,j])}$
represents a  tensor atom having  a ground atom  {\tt x} and  an index
{\tt [i,j]}.   It declares  a second-order tensor  (matrix) $\mmat{X}$
specified by {\tt x} whose index  $(i,j)$ is expressed by {\tt [i,j]}.
To  make  this correspondence  mathematically  rigorous,  we define  a
function $q_F(\cdot)$ from tensor atoms  to embedded tensors such that
$q_F({\tt tensor(x,[i,j])})=\mmat{X}_{i,j}$.  In  general, an index of
an $n$-order  tensor is  represented as an  $n$-length list  of Prolog
constants.   Note that  the  domain  of $q_F(.)$  can  be extended  to
programs  without  tensor atoms.   In  such  case, $q_F({\tt  x})$  is
defined  as one  (a scalar  value)  when a  ground atom  ${\tt x}$  is
logically true in Prolog. This treatment is consistent with PRISM in a
part of vanilla Prolog.

In  this paper,  as already  shown, we  use italic  fonts for  indices
appearing  in mathematical  expressions to  distinguish them  from the
corresponding ones appearing in a program.  So for a matrix $\mmat{X}$
specified by {\tt x}, we write an entry $\mmat{X}_{i,j}$ for the index
{\tt [i,j]}.  Also  to extract indices, we  define $\indexfunc_F$ such
that, for instance, $\indexfunc_F({\tt tensor(x,[i,j])})=(i,j)$.

In the  concrete numerical calculation, we  have to give the  range of
indices,  i.e.,  the size  of  embedded  tensors.  A  T-PRISM  program
contains a function $R({\tt i})$ that determines the range of an index
{\tt [i]} as  ${1, 2, ..., R({\tt i})}$.  Thus,  in the above example,
the  size  of  a  matrix $\mmat{X}_{i,j}$  is  determined  as  $R({\tt
i}) \times R({\tt j})$.

Once a T-PRISM program is given, a set of {\it tensor equation\/}s are
determined from  the program.   Consider a clause  having a  head $H$.
Let $H  \leftarrow W_{\ell} (\ell  =1,\ldots,L)$ be an  enumeration of
clauses   about  $H$   in  the   program.   For   each  $H$,   logical
equations\footnote{
Logical equivalence seems a more appropriate term, but  we use equation
for intuitiveness.
} of the  following form, which holds in the  least model of
the program, are derived.
\begin{align} \nonumber
H &\Leftrightarrow W_{1} \vee W_{2} \vee ... \vee W_{L} \\
W_{\ell} &\defeq B_{1\ell} \wedge B_{2\ell} \wedge ... \wedge B_{M_\ell \ell}
\label{eq:logic_eqs}
\end{align}
where $B_{*\ell}$ is a ground atom in the body $W_{\ell}$.

The  tensorized  semantics  is  obtained  by  expanding  the  mappings
$q_F(\cdot)$  and  $\indexfunc_F(\cdot)$   over  $F$  respectively  to
$q(\cdot)$ and  $\indexfunc(\cdot)$ for the entire  Herbrand base.  It
assigns various orders  of tensors to atoms as  their denotation. They
are inductively defined,  starting from tensor atoms,  by applying two
rules below to \prettyref{eq:logic_eqs}.
\begin{itemize}
\item {\it Disjunction rule}  \\
Given a disjunction of tensor  atoms, its embeddings
is defined as the summation of the tensor atoms, $A$ and $B$:
\[q(A \vee B) \defeq q(A)+ q(B).\]
% Here$\indexfunc(A_{a} \vee A_{b}) = \indexfunc(A_{a}) = \indexfunc(A_{b})$ is assumed.
%This rule say that an embedding of a disjunction consisting of vector-embedded atoms is a vector.
%When embedded values are multi-dimensional array, .
%Applying  this  rule  to  the  \prettyref{eq:t_prism_eqs},
%$\indexfunc(H)=\indexfunc(W_{1})=\indexfunc(W_{2})=...=\indexfunc(W_{L})$
%is derived.
%while
%assuming                         the                         condition
%$\indexfunc(W_{1})=\indexfunc(W_{2})=...=\indexfunc(W_{L})$ named {\it
%	disjunction                                            pre-condition}

%As condition, we introduce the following constraint:
%$\indexfunc(W_{1})=\indexfunc(W_{2})=...=\indexfunc(W_{L})$.

\item {\it Conjunction rule} \\
Given a conjunction of  tensor atoms with associated multi-dimensional
arrays for embedding, embeddings of  the conjunction is defined by the
{\em Einstein's notation\/} rule\footnote{
Although a  general formulation  of Einstein's  notation distinguishes
between  covariance and  contravariance,  indicated  by subscript  and
superscript indices  respectively, their distinction is  irrelevant to
our semantics, and hence ignored in this paper.
}. In the case where subscripts overlap  in the same term, the rule is
to take  a sum for that  subscript.  Note the non-overlap  indices are
not eliminated  by this  operation.  We call  the overlapping  index a
{\em dummy  index\/} whereas  a non-overlapping index  as a  {\em free
index\/}.  A set  of free indices is assigned  to $\indexfunc(A \wedge
B)$.      For      example,     put      $\indexfunc(A)=(i,j)$     and
$\indexfunc(B)=(j,k)$.  $\indexfunc(A \wedge  B)=(i,k)$ is computed by
this rule, and a matrix is embedded  to $q(A \wedge B)$.  As a result,
its  $i,k$-th element  $q(A  \wedge B)_{ik}$  is  computed as  $\sum_j
q(A)_{ij}  \cdot q(B)_{jk}$,  i.e.,  this  computation coincides  with
matrix multiplication.  When  the index is empty,  the resulting value
is  a  scalar (number)  and  the  value  of  a conjunction  of  scalar
embedding atoms is the product of their values.
\end{itemize}

By     applying    the     disjunction    and     conjunction    rules
to \prettyref{eq:logic_eqs}  recursively, we can derive  the following
tensor equations:
\begin{align}\nonumber
q(H) &= q(W_{1}) + q(W_{2}) + ... + q(W_{L}) \\
q(W_{\ell}) &= \einsum_{q,T}(B_{1\ell}, B_{2\ell}, ... , B_{M_\ell \ell})
\label{eq:tprism_eqs}
\end{align}
where  $\einsum_{q,T}$ is  a  sum-product operation  indicated by  the
Einstein's  notation in  the  conjunction rule.   The  derived set  of
tensor  equations  has  a  least solution  (as  all  coefficients  are
non-negative     and    all     derived    equations     have    upper
bounds)\cite{sato2017linear} and  we consider it as  the denotation of
the given T-PRISM program in the tensorized semantics.
In this  paper, we  assume \prettyref{eq:tprism_eqs}  is well-defined,
i.e., there is no mismatch of tensor dimensions.

%Suppose  the embedded  tensors are  scalar.   Then it  is possible  to
%define  categorical  distributions  using  $q_F(\cdot)$  and  some  of
%T-PRISM's operations  (as indicated by  a Markov chain program  in the
%section 4). So  the tensorized  semantics can  simulate the  PRISM
%semantics.   In this  case, the  disjunction  rule is  reduced to  the
%summation  of  probability of  exclusive  events  under the  exclusive
%assumption in PRISM \cite{Sato08a} and the conjunction rule is reduced
%to  the marginalization  of hidden  variables where  a dummy  index is
%regarded as  a hidden variable whose  range equals that of  the index.
%Hence, the new semantics can be  considered as a generalization of the
%distribution semantics.

Non-linear  operations   are  prerequisites  for  a   wider  range  of
applications including deep neural  networks.  To make them available,
T-PRISM provides  {\tt operator/1} indicating a  non-linear operation.
This built-in  predicate occurs in  an explanation graph  similarly to
{\tt tensor/2}.
The second equation in \prettyref{eq:tprism_eqs} is modified as follows:
\begin{align} \nonumber
q(W_{\ell}) = & op_1 \circ op_2 \circ ... \circ op_{O_{\ell}} \\
	&\circ   \einsum_{q,T}(B_{1\ell}, B_{2\ell}, ... , B_{M_\ell \ell}).
	\label{eq:tprism_eqsx} 
\end{align}
where  $op_1,  op_2,  ...,  op_{O_k}$ are  non-linear  operations,
$B_{*\ell}$ is a ground atom in the body $W_{\ell}$ except for  non-linear  operations, and
$\circ$  represents their  function  synthesis.   Note that  nonlinear
operations and  function synthesis are non-commutative.   For example,
let  $A$   and  $B$   be  tensor   atoms.   $B   \Leftrightarrow  {\tt
	operator}(f)  \wedge {\tt  operator}(g)  \wedge A$  is interpreted  as
$q(B) =  f \circ g  \circ q(A)$.  While ${\tt  operator}(f)\wedge {\tt
	operator}(g)   \wedge    A$   and   ${\tt    operator}(g)\wedge   {\tt
	operator}(f) \wedge A$ are  logically equivalent, they are interpreted
as different equations: $f(g(q(A)))$ and $g(f(q(B)))$, respectively.

\section{3. T-PRISM}

Theoretically T-PRISM is an implementation of the tensorized semantics
with a specific execution mechanism.  The main idea of T-PRISM program
execution is  that the  symbolic part  of program  execution including
recursion is processed by Prolog search and the numerical and parallel
part  such as  loop  for  sum-product computation  is  carried out  by
TensorFlow.
More  concretely  in  T-PRISM,  a  program is  first  compiled  to an {\em  explanation graph with
	Einstein's notation\/}, a set of logical equations
like \prettyref{eq:logic_eqs},
translated to a set of tensor equations
like  \prettyref{eq:tprism_eqs} and \prettyref{eq:tprism_eqsx}.
By removing  redundant repetitions,
Einstein's notation  enables sum-product computation much  faster than
that  in  PRISM  which  uses   sum  and  product  operations  naively.
Hereafter, we simply call it explanation graph.
This explanation graph is obtained by using   {\em  tabled search\/}  in  the
exhaustive search for all proofs for  the top-goal $G$ and by applying
dynamic  programming to  them \cite{Zhou08}.

% from the entire dataset by applying a gradient decent algorithm

To   train   parameters   in  embedded   tensors,   T-PRISM   supports
(stochastic) gradient decent algorithms  using   an   auto-differential
mechanism in TensorFlow.
We compute  the loss of  a dataset $D$ represented as  a set of  goals to 
learn parameters.
T-PRISM assumes the total loss is  defined to be the sum of individual
goal loss as follows:$Loss(D)=\sum_{G \in D} Loss(G)$.
%
%\begin{equation}
%Loss(D)=\sum_{G \in D} Loss(G).
%\end{equation}
%
Usually,  $Loss(G)$  is  a  function of  $q(G)$.   For  example,  when
parameter training uses  a negative log likelihood  loss function, the
goal loss function is defined as $Loss(G)=-\log q(G)$
where $q(G)$ stands for the likelihood of $G$.
For the  convenience  of  modeling,  the  T-PRISM  system  offers  related
built-ins for parameter training and  custom loss functions written by
Python programs with TensorFlow.
%A  loss function can be specified as a runtime argument for parameter training.
Utilizing this T-PRISM system based on tensorized semantics with non-linear operators,
various models including matrix computation, deep neural networks, and the models written by recursive programs can be designed.

\section{4. T-PRISM programs for knowledge graphs}

This section explains sample programs of T-PRISM through knowledge
graph modeling using DistMult  \cite{Yang15}, After that,  we present
learning experiments  with real  datasets: the
FB15k                             and                             WN18
datasets \cite{bordes2014semantic,bordes2013translating}.

The  T-PRISM system  provides a  special predicate  {\tt tensor/2} to define $T_F$ and $q_F$.  Also it
provides  {\tt  set\_index\_range/2}  to   define  $R$.   Using  these
built-in  predicates, T-PRISM  programs are  written just  like Prolog
programs.

In addition  to these  special predicates, our preliminary implementation
requires  to declare  all  tensor  atoms used  in  a  program using  a
built-in predicates {\tt index\_list/2}.  The first argument specifies
a  tensor atom,  and the  second one  specifies a  list of  the second
arguments  of {\tt  tensor/2}s  that occur  during program  execution,
i.e., its possible  use of indices for computing a cost function.
%In
%this preliminary system,  users need to understand the  type of tensor
%computation  in  the  program  and manually  write  appropriate  index
%declarations.   We  expect  that  this problem  will  be  improved  by
%automated inference in the future.

\begin{figure}[tb]
	%\rule{24em}{0.10mm}\\[0.6em]
	\rule{0.45\textwidth}{0.10mm}\\ [-1em]
	\begin{verbatim}
 1: index_list(v(_),[[i]]).
 2: index_list(r(_),[[i]]).
 3: :-set_index_range(i,20).
 4:
 5: rel(S,R,O):-
 6:   tensor(v(S),[i]),
 7:   tensor(v(O),[i]),
 8:   tensor(r(R),[i]).
	\end{verbatim}
\rule{0.45\textwidth}{0.10mm}\\ [-1.5em]
\caption{simple DistMult program in T-PRISM}
\label{fig:tprism-distmult}
\end{figure}

We  first look  at  a simple T-PRISM program  used  for  link prediction  in
knowledge  graphs.  A  knowledge graph  consists of  triples (subject,
relation, object), represented  by a  ground atom  {\tt rel}(subject,  relation, object) using the {\tt  rel}/3 predicate.
So logically  speaking, a knowledge
graph is nothing but a set of ground atoms.

For  link   prediction  in   knowledge  graphs,  we   choose  DistMult
model \cite{Yang15}, one  of the standard knowledge  graph models, for
its simplicity.  In DistMult, entities $s$, $o$, $r$ ($s$ for subject,
$o$ for  object, $r$ for  relation) are encoded as  N-dimensional real
vectors, $\mvec{s}$, $\mvec{o}$, and  $\mvec{r}$ respectively, and the
prediction is made based on the score of $(s,r,o)$ computed by:
\begin{equation}
f(s,r,o)=\sum_i \mvec{s}_i \mvec{r}_i \mvec{o}_i.
\label{eq:distmult}
\end{equation}

In      T-PRISM,      a       DistMult      model      is      written
like   \prettyref{fig:tprism-distmult}.     The   lines   1    and   2
introduce \verb|v(_)| and \verb|r(_)|.  They are embedded vectors with
an index  \verb|i|. The lines  3 says the  range of index  \verb|i| is
$0  \leq $\verb|i|$<  20$.  The  lines 5-8  represent an  equation for
sum-product computation  for the  three embedded vectors  required for
computing the scores in \prettyref{eq:distmult}.

\begin{figure}[tb]
	%\rule{24em}{0.10mm}\\[0.6em]
	\rule{0.45\textwidth}{0.10mm}\\ [-1em]
	\begin{verbatim}
 1: index_list(v(_),[[i]]).
 2: index_list(v,[[o,i]]).
 3: index_list(r(_),[[i]]).
 4: :-set_index_range(i,256).
 5: :-set_index_range(o,14951).
 6: 
 7: rel(S,R,O):-
 8:   tensor(v(S),[i]),
 9:   tensor(v,[o,i]),
10:   tensor(r(R),[i]).
	\end{verbatim}
	\rule{0.45\textwidth}{0.10mm}\\ [-1.5em]
	\caption{DistMult program for large-scale data}
	\label{fig:distmult_kb}
\end{figure}
\begin{table}
	\begin{center}
	\begin{tabular}{|c|c|c|}
	\hline 
	& FB15K & WN18 \\ 
	\hline 
	MRR	&  54.16\% & 60.62\% \\ 
	\hline 
	HIT@10	& 75.88\% &  86.00\% \\ 
	\hline 
	HIT@1	&  42.22\%& 47.04\% \\ 
	\hline 
	\end{tabular}
	\end{center}
	\caption{DistMult (\prettyref{fig:distmult_kb}) performance using T-PRISM}
	\label{tbl:result_distmult}
\end{table}

Next, we apply T-PRISM to  link prediction in large  size  datasets.
For more efficient computation, we rewrite  a  DistMult  program  as
in    \prettyref{fig:distmult_kb}.    This    program   outputs    the
$o$-dimensional vector, and  computes scores for all  entities at once
(the  number of  entities in  FB15k  is 14951). Furthermore, we use a mini-batch method, an essential method  in  deep  learning  and  other optimization  problems  from  the  viewpoint  of  learning  speed  and performance \cite{Goodfellow-et-al-2016}.
Note that {\tt O} is  a singleton  variable, i.e.,  not used  in the  Prolog search  and construction  of  tensor equations;  however,  it  is instantiated  as
labels and required to compute a  loss function described
as below.

For our link prediction experiment, we employ a loss function for the knowledge graph as follows:
\[Loss(rel(S,R,O)) = \left[ f(S, R, O) - f(S, R, O') -\gamma \right]_{+}\]
where $[\cdot]_{+}$ means  a hinge loss $\max(0,\cdot)$  and a negative
sample  $O'$ is  uniformly sampled  from all  entities.  We  use this
hinge  loss  function  with   $L2$  regularization  and  its  learning
parameters set to $\lambda=1.0e-5$ and $\gamma=1$.

In  this experiment,  we use  two standard  datasets:
WN18  from      WordNet       and
FB15k  from  the  Freebase.  The  purpose
here is not to outperform known models for knowledge graphs but rather
to demonstrate that  T-PRISM can implement them very  compactly with a
declarative program.

For FB15K,  we use the following  settings: the batch size  $M = 512$,
the number  of negative  sampling for one  positive sample  $N_{neg} =
100$   and   select   as   a   learning   rate   optimization   method
Adam \cite{kingma2014adam},  with the  initial  learning rate  $0.001$.
For WN18,  we set $M$ to  $256$ and other  setting is the same  as the
case of FB15K.

This experiment  shows that while  the DistMult model  written in
PRISM \cite{kojima2018ijar} is incapable  of dealing with these datasets due  to their sheer
data size,  the T-PRISM  implementation of  a DistMult model  can handle
them  and  achieves  reasonable  prediction  performance  as  listed
in \prettyref{tbl:result_distmult}.   Thanks to a GPU,
this experiment is finished in 5 hours\footnote{
NVIDIA Tesla P40 GPU was used.
}.

%eval/20180902_0425/fb15ko.txt
%MRR: 0.4985546516132536
%HIT@10: 0.6150395287027476
%HIT@1: 0.415618492

%time:532.0558030605316[sec]
%: step 26 train loss: 1.2367665194619806 valid loss: 1.2060938786785558
%traing time:14397.300181865692[sec]

%a recursive programs for
%a  finite-state discrete-time  Markov chain,  and transitive  closure
%computation with  cyclic explanation  graphs.  

\section{Conclusion}

We proposed  an innovative tensorized logic  language T-PRISM together
with its  tensorized semantics,  enabling declarative  tensor modeling
for a large-scale  data.  We explained how programs  are compiled into
tensor equations by way of explanation graphs with Einstein's notation
using Prolog's  tabled search, and executed  on TensorFlow efficiently.
T-PRISM  supports a rich  array of
non-linear  operations   and  custom  cost  functions   for  parameter
training. Using  our preliminary T-PRISM system,  we also demonstrated
the  modeling of  knowledge  graphs.   Future work  includes
applying T-PRISM to real world  problems and showing the effectiveness
of logic-based declarative modeling.
%,  transitive closure  computation, discrete  Markov  chain and  neural  networks

%We are
%planning to release T-PRISM as open source software
%\footnote{
	%
	%\url{TBD:URL}
	%
%}.

\if 0
\section*{Acknowledgments}
This research  is supported in part by  a project commissioned
by the  New Energy and Industrial  Technology Development Organization
(NEDO).
\fi
\small
\bibliographystyle{aaai}
\bibliography{./aaai2019}

\end{document}